\documentclass[sigconf]{acmart}
\AtBeginDocument{%
  \providecommand\BibTeX{{%
    \normalfont B\kern-0.5em{\scshape i\kern-0.25em b}\kern-0.8em\TeX}}}

\copyrightyear{2022}
\acmYear{2022}
\setcopyright{rightsretained}
\acmConference[MADiMa '22]{Proceedings of the 7th International
Workshop on Multimedia Assisted Dietary Management}{October 10,
2022}{Lisboa, Portugal}
\acmBooktitle{Proceedings of the 7th International Workshop on
Multimedia Assisted Dietary Management (MADiMa '22), October 10,
2022, Lisboa, Portugal}
\acmDOI{10.1145/3552484.3555747}
\acmISBN{978-1-4503-9502-1/22/10}
\newcommand\sbullet[1][.5]{\mathbin{\vcenter{\hbox{\scalebox{#1}{$\bullet$}}}}}

\usepackage{algorithm}
\usepackage{amsmath}%%,amssymb}
\usepackage{bm}
\begin{document}
\title{SIMULATING PERSONAL FOOD CONSUMPTION PATTERNS USING A MODIFIED MARKOV CHAIN}

\author{Xinyue Pan}
%%\orcid{1234-5678-9012}
\affiliation{%
  \institution{Purdue University}
  \streetaddress{501 Northwestern Ave}
  \city{West Lafayette}
  \state{Indiana}
  \country{United States}
  \postcode{47906-2035}}
\email{pan161@purdue.edu}

\author{Jiangpeng He}
\affiliation{%
  \institution{Purdue University}
  \streetaddress{501 Northwestern Ave}
  \city{West Lafayette}
  \country{United States}
  \postcode{47906-2035}}
\email{he416@purdue.edu}

\author{Andrew Peng}
\affiliation{%
  \institution{Purdue University}
  \streetaddress{501 Northwestern Ave}
  \city{West Lafayette}
  \country{United States}
  \postcode{47906-2035}}
\email{andrew.w.peng@gmail.com}

\author{Fengqing Zhu}
\affiliation{%
  \institution{Purdue University}
  \streetaddress{501 Northwestern Ave}
  \city{West Lafayette}
  \country{United States}
  \postcode{47906-2035}}
\email{zhu0@purdue.edu}
\renewcommand{\shortauthors}{}

\begin{abstract}
Food image classification serves as the foundation of image-based dietary assessment to predict food categories. Since there are many different food classes in real life, conventional models cannot achieve sufficiently high accuracy. Personalized classifiers aim to largely improve the accuracy of food image classification for each individual. However, a lack of public personal food consumption data proves to be a challenge for training such models. To address this issue, we propose a novel framework to simulate personal food consumption data patterns, leveraging the use of a modified Markov chain model and self-supervised learning. Our method is capable of creating an accurate future data pattern from a limited amount of initial data, and our simulated data patterns can be closely correlated with the initial data pattern. Furthermore, we use Dynamic Time Warping distance and Kullback-Leibler divergence as metrics to evaluate the effectiveness of our method on the public Food-101 dataset. Our experimental results demonstrate promising performance compared with random simulation and the original Markov chain method.

\end{abstract}

%%
%% The code below is simulated by the tool at http://dl.acm.org/ccs.cfm.




%%  \includegraphics[width=\textwidth]{sampleteaser}
%%  \caption{Seattle Mariners at Spring Training, 2010.}
%%  \Description{Enjoying the baseball game from the third-base
%%  seats. Ichiro Suzuki preparing to bat.}

\maketitle

\section{Introduction}
Image-based methods have been developed to provide timely feedback on an individual's dietary intake~\cite{shao2021_ibda} with reduced user effort compared to traditional self-reported methods \cite{boushey-pns2017}. 
% to monitor the health status of a person by providing the nutrition and energy intake based on eating images. 
Classification of food images is typically the first and most fundamental step in automated image-based food analysis~\cite{he2020multitask, he2021end}. 
% Food classification serves as the foundation of image-based dietary assessment to predict the food categories. 
Most existing works focus on designing methods to improve the accuracy of food classification using static food image datasets~\cite{shao21, Phiphiphatphaisit20,ma21, McAllister18, mao2020visual, mao2021visual2}. However, static datasets such as Food-101\cite{bossard14} or VireoFood-172\cite{chen16} are limited to training fixed classifiers, which may not be suitable for real-life scenarios because each person has their unique food consumption patterns. In addition, accurate classification of food images is challenging due to the intra-class diversity and inter-class similarity. Food images may have diverse appearances for the same food class due to different cooking styles, and different food classes may have a similar visual appearance. 
% {As shown in \cite{shota18}, using Convolution Neural Networks (CNNs) to classify food images in real life scenarios only obtains an accuracy of around 20\% because there are food classes that has not been appeared in fixed dataset. }
To improve the accuracy of food image classification and tailor it to an individual food consumption pattern, certain researchers have designed personalized classifiers~\cite{shota18}. Instead of classifying images on static datasets, a personalized classifier uses food images based on a personal food consumption pattern. 
% A personalized classifier could address the intra-class diversity and inter-class similarity problems that occur in the fixed class classifier. In food images, intra-class diversity means different cooking styles in the same food class, and inter-class similarity means similar food images across different classes. These problems can cause the confusion in classification of food images in the fixed class classifier.  
%Moreover, food classes in the real world are different from currently available static datasets. For example, a static dataset contains fish, chicken, and egg classes while in the real world a person eats beef, breast, and broccoli. A personalized classifier can address this issue by classifying images over the personal eating sequence so that it knows what kinds of food a person eats and the pattern of a person's diet.

Currently, there are few works that focus on designing a personalized classifier for food images. One of the main challenges is the lack of publicly available personal eating datasets. Q. Yu \textit{et al.} designed a personalized classifier by building a personal database for each person incrementally and comparing the similarity of images at each time step with images in each food class, combined with a time-dependent model to predict the food image at each time step \cite{qing18}. However, it took around 2 years to collect sufficient eating data from each person to train their model, which is time-consuming and difficult to generalize.
% reproduce since the data is not public. 
% Their dataset contains eating data from 20,000 user with 300 records for each person and a total of 276,000 images are collected. 
Therefore, how to efficiently simulate personal food eating data patterns remains an open problem. A simulated data pattern that can closely mimic real life scenario would be essential for developing personalized food classification because the simulated data pattern can help train and test the personalized classifier without collecting large amounts of eating data from different people. In this paper, we focus on simulating food consumption patterns that correlates well with the initial data pattern and can be used for training and testing a personalized classifier. 
% In order to make the simulated data pattern useful, we have to make sure the simulated pattern allows a machine learning framework to learn. As said by \cite{shota18}, random data cannot help a personalized classifier learn. Therefore, we have to make the simulated pattern possesses some correlations inside. 
To our best knowledge, we are the first to simulate food consumption patterns that can be used for training personalized classifiers.

In this work, we propose a novel method to simulate personal food consumption patterns. Our method builds a food consumption data pattern of any length from an initial data pattern, which generally consists of one to two weeks of eating data. Compared to collecting real-life personal data, our method is much more efficient while allowing the simulated data pattern to correlate well with the initial data pattern. Specifically, we leverage a Markov chain model to predict the occurrence of food in the future. The Markov chain model has the advantage of not requiring a large amount of data to train. Moreover, the logic behind the Markov chain model is applicable in our scenario because it uses conditional probability given what foods were eaten in the past, we simulate what will be eaten in the future. We modified the original model to allow more flexibility in predicting what a person tends to eat to mimic real-life scenarios. 

% Markov chains are a stochastic process that predict future events, and it has been applied in natural language processing to simulate sentences, music, etc. We use the Markov chain model because it does not require a large amount of data to train and it does not include a complex probabilistic model. In addition, our scenario is similar to the applications mentioned above. However, the original Markov chain method is intended to predict the food that a person tends to eat in the end. This phenomenon is called convergence and this is not realistic. Our goal is to simulate a pattern that can have some fluctuations in diets, so we need to make some modifications to the original Markov chain method.    

% based on our multiple experiments, after many time units' inference, the model will give repetitive predictions so that the simulated food sequence is not close to real life. We have made some modifications to make the simulated food sequence close to real life. Our modified Markov chain method works well after some evaluation based on the Dynamic timing distance between the simulated food sequence and the initial food sequence. 

The contribution of our work is summarized as follows.
\begin{itemize}
    \item We modify the Markov chain model to simulate food consumption patterns that can be used for personalized classifiers, considering the case of eating foods not appeared in the initially provided pattern. 
    % (Section \ref{sec:dps} and \ref{sec:new_class})
    \item We propose the use of Dynamic Time Warping distance and Kullback-Leibler divergence to show the success of simulated food consumption patterns.
    % (Section \ref{sec:experiments})
    \item We sample images for the food consumption pattern by building a normal distribution based on the visual similarity clustering for each class and personal preference.
    % (Section \ref{sec:sampling})
\end{itemize}
%%Designing a personalized classifier has some challenges. Although \cite{qing18}\cite{shota18} has collected two years of food eating data patterns for each person, such datasets are not open to the public. In addition, there are cold-start and new class learning problems when designing a personalized classifier.

%%To make the simulation close to real-world scenarios from other aspects, when simulating eating data patterns, we have also defined new classes when extending the initially provided data pattern since a person might want to eat new kinds of food after eating many old foods. Different cooking styles in the same food class have also been considered when simulating the data pattern since a person might prefer some specific cooking styles in a single food class. More details will be discussed in section3.

% Our paper is organized as follows: Section~\ref{sec:related work} will discuss other works that are similar to ours. Section~\ref{sec:methods for simulating eating data pattern} will describe the technical details of simulating eating data patterns. Section~\ref{sec:experiments} will present our experimental results and discuss our way of simulating the food eating pattern.

\section{Related Work}
\label{sec:related work}
\subsection{Food Image Classification}
Many different models of food image classification have already been published. P. Ma \textit{et al.} compiled a food image dataset named ChinaMartFood-109 with nutrition information \cite{ma21}. They tried different network architectures on this dataset, such as VGG~\cite{karen14}, ResNet~\cite{he15}, Wide ResNet~\cite{zagoruyko16} , and InceptionV3~\cite{szegedy15}, to train the classifier and compare the classification accuracy. They found that InceptionV3 obtained the best food recognition accuracy. In another work~\cite{McAllister18}, P. McAllister \textit{et al.} found that Resnet-152 features provide better generalization for food image classification on popular datasets such as Food 5K, Food-11, RawFooT-DB~\cite{Cusano_2015}, and Food-101~\cite{bossard14}. S. Phiphiphatphaisit and O. Surinta \cite{Phiphiphatphaisit20} also applied modified MobileNet architecture to improve the food image classification accuracy and reduce computational time. Besides, recently the food image classification has been studied under a more realistic continual learning scenario where new foods come sequentially overtime~\cite{ILIO, He_2021_ICCVW, he2022_icip}. However, none of existing work can tailor to individual food consumption patterns, which can help provide more accurate dietary assessment results. 

\subsection{Personalized Classifier}
S. Horiguch \textit{et al.} proposed a dataset named FLD that compiled 1.5 million images of eating data from 20,000 people over two years~\cite{shota18} . It trained a fixed class classifier on static food datasets first and then applied the classifier to the collected eating data to extract features for each image at each time step. Using the nearest class mean method, a personalized classifier can be constructed. Q. Yu \textit{et al.} improved the classification method that was originally proposed in \cite{shota18} by applying a time-dependent food distribution model~\cite{qing18}. However, the data collection is time-consuming and the dataset is not open to the public.

\section{System Overview}
\label{sec:system}
\begin{figure}[h]
    \centering
    \includegraphics[width=8.6cm, height=3.1cm]{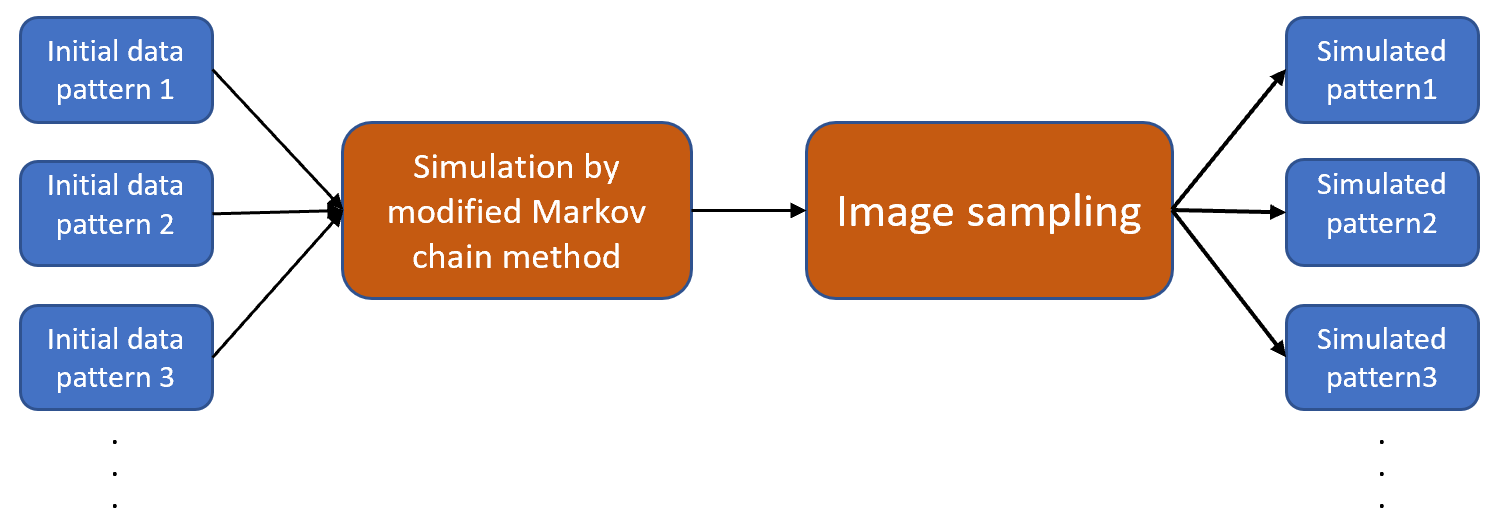}
    \caption{System overview of the simulation}
    \label{fig:sys_overview}
\end{figure}

The objective of this work is to simulate personalized food consumption patterns by using (1) initial data pattern, and (2) static food image datasets as the input. The initial data pattern is provided by the user, which contains what a person eats in the past few days. Allowing the user to provide a short initial data pattern ensures that the simulation is not done randomly and there are some correlations within the simulated pattern. In addition, it is easier to collect short-term food records from each person than a long-term ones. The overview of our proposed food consumption pattern simulation system is shown in Figure \ref{fig:sys_overview}, which includes simulation based on modified Markov chain method that will be described in Section\ref{sec:modified markov chain} and the image sampling in Section \ref{sec:sampling}. The output of our system is the personalized food consumption pattern that simulates what a person eats every day.
% This section will present a holistic overview of the system which includes the introduced components. Figure \ref{fig:sys_overview} describes the overview of our simulation system. Users input what they ate in the last few days as the initial data pattern. The length of the initial data pattern can also be varied. Different initial data patterns will go through the process of simulation and image sampling to produce complete simulated patterns as output.

The simulation using the modified Markov chain method is introduced in section \ref{sec:dps}. The purpose is to simulate a food consumption pattern that has some correlations with the initial data pattern so that the simulated data pattern is learnable by the personalized classifier. Since a person may eat new types of food from time to time, we can choose to incorporate new food classes that have not appeared in the initial data pattern, which is described in Section \ref{sec:new_class}. 

We introduce the image sampling process in Section \ref{sec:sampling}. After the simulated data pattern is generated, which contains food types that a person may eat, we need to sample food images for each food type to complete the food image data simulation. 
% Because after simulation using modified Markov chain method, the simulated data pattern only contains food types that a person might want to eat but a complete simulation should be image-based. Therefore, we need to sample food images for each food type that has appeared in the simulated data pattern to make the simulation complete. 
We leverage visual similarity clustering and cluster sampling, which are described in Section \ref{sec:vis_clus} and section \ref{sec:clu_smp}, respectively. The purpose is to mimic the scenario that a person typically prefers certain cooking styles for each type of food so that the simulated food consumption pattern is more realistic.

\section{Methods For Simulating Food Consumption Pattern}
\label{sec:methods for simulating eating data pattern}
\begin{figure}[htp]
    \label{sec:pipeline}
    \centering
    \includegraphics[width=6.8cm,height=7.8cm]{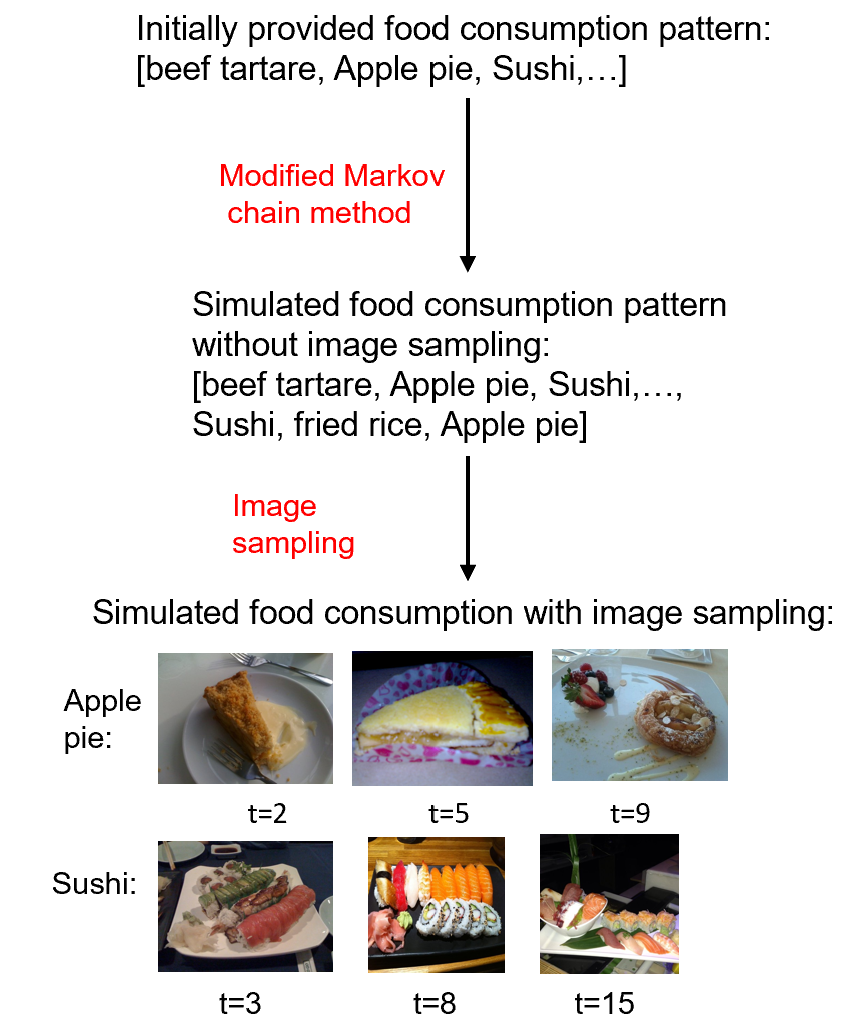}
    \caption{Pipeline for simulating an eating data pattern}
    \label{fig:pipeline}
\end{figure}

We propose a new framework to simulate food consumption patterns of any length that closely mimics a provided initial data pattern while allowing flexibility for generalization. We assume an initial food consumption pattern is available for each individual. We then extend this data pattern by building a modified Markov chain model, which does not require a large amount of training data. A Markov chain model is suitable in our scenario since the simulation of the data pattern is based on what was eaten in the past. However, the original Markov chain method can result large difference in the probability of food appearing between the simulated data pattern and the initial data pattern. This is because it tends to forget the probability distribution of the original data pattern in the later part of the simulation. Therefore, we propose a modified Markov chain model. We also sample images for food classes that appeared in food consumption patterns via a clustering method. Figure \ref{fig:pipeline} describes the overall pipeline of our method. The user first provides an initial food consumption pattern, then we simulate the food consumption pattern by extending this initial data pattern using a modified Markov chain method. Finally, we sample images of each food class that appeared in the simulated data pattern.
%%\begin{comment}

%%\end{comment}

\subsection{Modified Markov Chain Method to Simulate food consumption Pattern}
\label{sec:modified markov chain}

\subsubsection{Original Markov Chain Method}
\label{sec:original}
T. Almutiri and F. Nadeem presented a survey of example applications that use the Markov chain method~\cite{almutiri22}. The Markov chain method \cite{richard} uses conditional probability to simulate the next state from only the previous state. It can be represented in the following equation ~\ref{eq:prob}

\begin{equation}
P(S_t=A|S_1...S_{t-1}) = P(S_t=A|S_{t-1}),
\label{eq:prob}
\end{equation}
where $S_t$ represents the predicted state at time $t$ and $S_{t-1}$ represents the current state at time $t-1$. There is also a decision array that represents the probability of each state that can happen in the current time, as shown in equation ~\ref{eq:dec_arr}

\begin{equation}
P_{decision} = [P_A, P_B, P_C, P_D, ...],
\label{eq:dec_arr}
\end{equation}
where $P_A$ represents the probability of state A in the current unit of time and the same for $P_B$, $P_C$, etc. To predict the probability of the next state, we also need a transition matrix, which represents the conditional probability of transitioning from one state to another in different combinations:

\begin{equation}
P_{trans} = 
\begin{bmatrix}
P_{A|A} & P_{B|A} & ... \\
P_{A|B} & P_{B|B} & ... \\
... & ... & ...
\end{bmatrix},	
\end{equation}
where $P_{A|B}$ represents the probability of transitioning from state B to state A. Each row will be normalized such that the summation of each row equals 1. Then, the probability of predicting different states at the next unit of time is calculated using equation ~\ref{eq:dec_arr2}

\begin{equation}
P_{new\_decision} = P_{old\_decision} * P_{trans}.
\label{eq:dec_arr2}
\end{equation}
The highest probability element in the decision array is then selected as the prediction. 

\subsubsection{Food consumption Pattern Simulation using modified Markov chain method}
\label{sec:dps}
\begin{figure*}[h]
    \label{sec:markov}
    \centering
    \includegraphics[width=17cm,height=3.3cm]{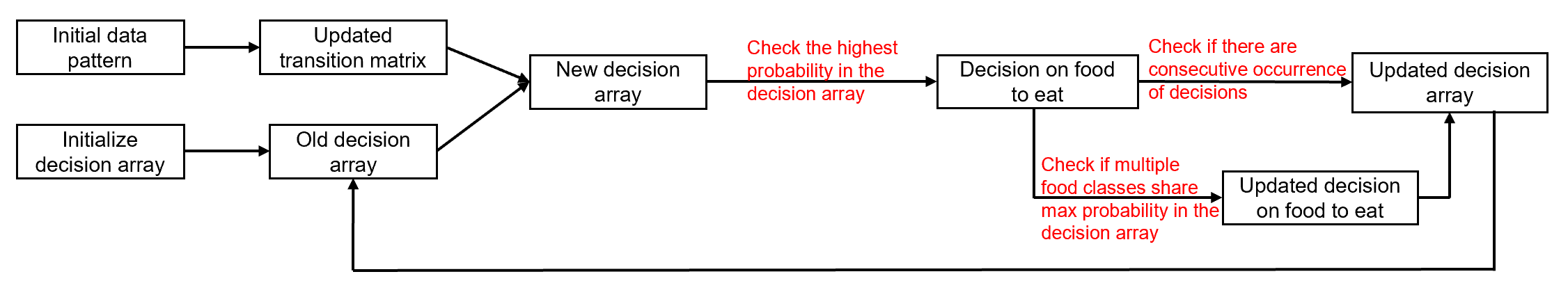}
    \caption{Modified Markov chain method for simulating data pattern}
    \label{fig:modified markov chain}
\end{figure*}

Figure ~\ref{fig:modified markov chain} shows our modified Markov chain method for simulating food consumption patterns. The old decision array represents the probability of each food class in the last time step. The new decision array represents the probability of each food class in the current time step after multiplying the old decision array with the transition matrix. The updated decision array represents the updated probability of each food class in the current step after handling special circumstances during the simulation of food consumption patterns. Given an initial food consumption pattern, we use the Markov chain method to simulate a subsequent data pattern. However, the original Markov chain method results in repetitive occurrences of the same food class, which is unrealistic. 
% after some experiments, we found that the resulting data pattern always has a repetitive pattern, which is not realistic. Although such a case is possible, it is rare. 
This is expected because the goal of a Markov chain method is to predict the state a data pattern converges to. We also noticed a few other issues with the Markov chain method. For example, if multiple food classes could have the same highest probability, the first one is always chosen by default. There are also sometimes zero rows in the transition matrix if the last food class in the initial data pattern only appears once, which means that there is no data to construct conditional probability for this particular class. To address these issues, we modified the original Markov chain method to lower the probability of simulating repetitive data patterns and make decisions less biased. Our modifications include:
\begin{itemize}
    \item If there are repetitive occurrences of one food class when simulating the food consumption pattern, we reinitialize the decision array randomly by assigning a food class with a probability of 1 in the decision array.
    \item If two or more food classes share the highest probability, we randomly select a food class from the classes sharing the highest probability and reinitialize the decision array.
    \item For any zero rows in the transition matrix, we replace the zero row with the average probabilities calculated from other non-zero rows.
\end{itemize}

%Figure \ref{fig:modified markov chain} shows the pipeline of our modified Markov chain method.

% In addition, there are sometimes more than one food classes with same highest probability, and computer programs always pick up the first argument that obtains the highest probability. We also found that there are zero rows in the transition matrix because the last food item in the initial data pattern could only appear once, so there is no data to construct conditional probability for this particular class. Therefore, to bypass such restrictions, we needed to make some modifications to the original Markov Chain method to lower the probability of simulating repetitive data patterns and make decisions without bias. The modifications we made are as follows:\begin{itemize}
% \item 3.1.1 If, in the last 2 units of time, there is only one food class appearing, reinitialize the decision array with values of one in random food class and zero in other food classes.

% \item 3.1.2 If two or more food classes have an equal highest probability, randomly select a food class and reinitialize the decision array.

% \item 3.1.3 If there is any row full of zeros in the transition matrix, fill that row using the average probabilities calculated from other rows.
% \end{itemize}

\subsection{Incorporating New food Classes}
\label{sec:new_class}
It is highly probable that the initial food consumption pattern does not include all the foods a person may eat, or that new foods may be consumed over time. Therefore, our method incorporates new food classes as follows.

\subsubsection{Adding New Food Class}
We first need to decide how to define a new food class. At each time step, there is a probability that new food is consumed. Therefore, we build a probability model to estimate the likelihood of eating a new food at each time step, as represented in equation ~\ref{eq:prob_new}

\begin{equation}
    P_{new} = \frac{1}{c_{new} + 1} * (\frac{1}{c_{total} + 1})^{\frac{1}{x_t - x_{t-1}+1}}
\label{eq:prob_new}
\end{equation}
where $c_{new}$ represents the total number of new classes defined after the initially provided data pattern, $c_{total}$ represents the total number of food classes in the data pattern, $x_t$ represents the current time index, and $x_{t-1}$ represents the last time index when a new food class is defined. The idea behind this equation is that the longer a person eats the old foods since the last time they ate new food, the more likely they are to eat another new food, and vice versa. If the probability of adding a new food class is larger than the probability of eating an existing food, a new food class will be added to the data pattern instead of inferring what food to eat from the modified Markov chain method.

\subsubsection{Expansion Of Transition Matrix And Decision Array}
When we define a new food class, we add one row and one column to the transition matrix. The simplest way to fill the additional row or column is to assign random values to each element. However, this could lead to a higher probability for the newly added entries during the update of the decision array. When a decision array multiplies with the transition matrix, each element in the decision matrix corresponds to the dot product between the decision array and the corresponding column in the transition matrix. If the corresponding column does not have any sparsity (defined as the proportion of zeroes), the element will have a larger value, which means a higher probability to be selected in the updated decision array. To address this issue, we randomly add sparsity in the additional column to keep the sparsity close to that of the original transition matrix. This would ensure that the decision at each time step is always alternating between new food classes and existing food classes.

\subsection{Image Sampling For simulated Food consumption Pattern}
\label{sec:sampling}
After obtaining the food consumption pattern, we can sample images from existing food datasets such as Food-101~\cite{bossard14} to simulate personalized consumption data. As indicated in~\cite{qing18,shota18}, one of the characteristics of personalized data is that foods from the same class are more visually similar, as each person may prefer certain cooking styles for the same foods. This motivates us to perform visual similarity clustering within each food class first, and then sample images in the same cluster to make the data pattern more realistic. 

\begin{figure}[htp]
    \label{sec:clustering}
    \centering
    \includegraphics[width=5.2cm, height=9cm]{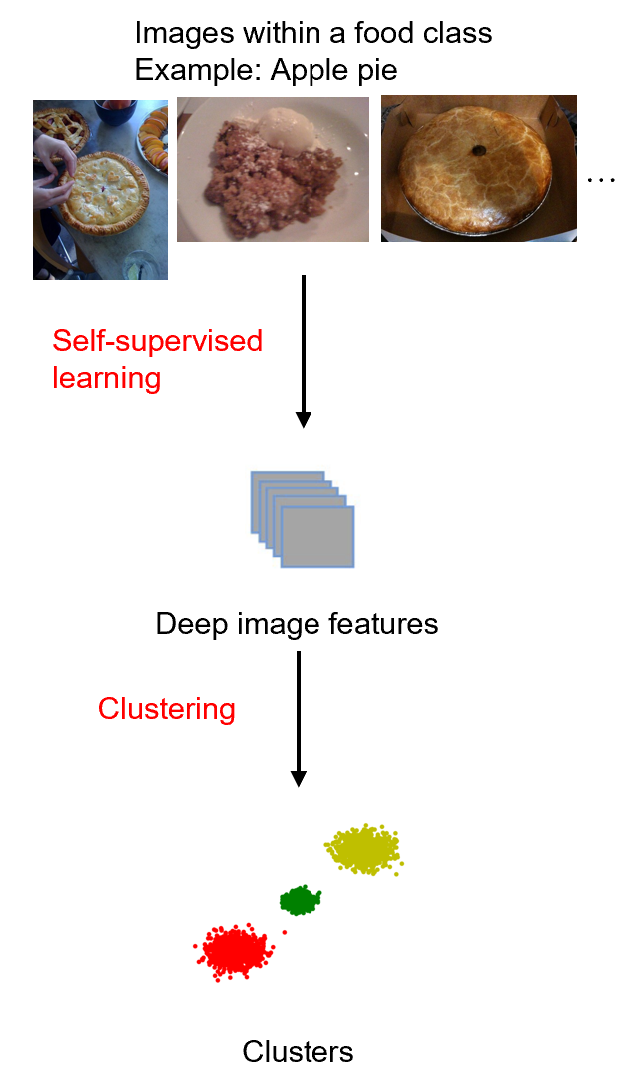}
    \caption{Pipeline for visual similarity clustering}
    \label{fig:clustering}
\end{figure}

\subsubsection{Visual Similarity Clustering}
\label{sec:vis_clus}
To cluster visually similar images within each food class, we first need to learn the discriminative features for each food image. Figure \ref{fig:clustering} shows the pipeline for visual similarity clustering. The first step is performing image feature extraction and the second step is clustering images based on these extracted features. 
To extract discriminative image features, we propose to apply self-supervised learning techniques to learn the visual representation of food images with static datasets. Our pipeline can work with any existing self-supervised approaches. In this work, we apply SimSiam \cite{chen21} as an example to illustrate our method.

\textbf{SimSiam} \cite{chen21} learns the visual representation by passing two different augmented views of an input image into an encoder network and through a projection MLP head. Finally, it will pass through the prediction MLP to maximize the agreement with the other augmentation. SimSiam is trained on minimizing negative cosine similarity between the prediction output $p(\cdot)$ on one path and a stop gradient of the projection output $stop\_grad(h(\cdot))$ on the other. The same process will be repeated for the other pair of paths. The two different negative cosine similarities sum together to form the total loss. Here, the stop gradient is a critical part to avoid a collapsing solution. The whole process can be formulated as equation ~\ref{eq:loss}. 
\begin{equation}
    Loss = \frac{1}{2} D(p_1(\cdot),stop\_grad(h_2(\cdot)) + \frac{1}{2} D(p_2(\cdot),stop\_grad(h_1(\cdot))
\label{eq:loss}
\end{equation}
where D refers to the negative cosine similarity between two different vectors.

% experiment with two methods. One is pre-trained convnet\cite{caron18} and the other is self Contrastive learning\cite{chen20} to see the comparison. Both ways of feature extraction are based on unsupervised learning. Pre-trained convnet\cite{caron18} is based on the features trained in the CNN model while self Contrastive learning\cite{chen20} is to maximize the agreement between two features that come from the same image but different data augmentation.

\textbf{Clustering:} After self-supervised learning, we apply the Power Iteration Clustering (PIC) \cite{lin10} as our clustering approach, which is a graph based method. One advantage of PIC is that the number of simulated clusters is not predefined, so there will be more clusters if the food class has higher intra-class variations and vice versa. Given $n_c$ images for one food class $c$, we first simulate the nearest neighbor graph by connecting their $10$ neighbor data points in the Euclidean space using extracted feature embeddings. Let $f(\textbf{x}_i)$ denote the extracted feature for the $i$-th image. The sparse graph matrix $G = \mathbb{R}^{n_c\times n_c}$ with zeros on the diagonal and the remaining elements of $G$ are defined by equation ~\ref{eq:matrix}
\begin{equation}
e_{i,j} = exp^-\frac{||f(\textbf{x}_i) - f(\textbf{x}_j)||^2}{\sigma^2} 
\label{eq:matrix}
\end{equation}
where $\sigma$ denotes the bandwidth parameter and we empirically use $\sigma = 0.5$ in this work. Then, we initialize a starting vector $s^{n_c\times1} = [\frac{1}{n_c}, ..., \frac{1}{n_c}]^T$ and iteratively update it using Equation~\ref{eq:pic}
\begin{equation} \label{eq:pic}
\begin{aligned}
s = L_1(\alpha(G + G^t)s + (1-\alpha)s)
\end{aligned}
\end{equation}
where $\alpha = 0.001$ refers to a regularization parameter and $L_1(\sbullet[.5])$ denotes the L-1 normalization step. The simulated clusters are given by the connected components of a directed, unweighted subgraph of $G$ denoted as $\tilde{G}$. We set $\tilde{G}_{i,j} = 1$ if $j = \textit{argmax}_je_{i,j}(s_j - s_i)$ where $s_i$ refers to the $i$-th element of the vector. Note that no edge starts from $i$ if $\{\forall j\neq i, s_j \leq s_i\}$, \textit{i.e. }$s_i$ is a local maximum.

\subsubsection{Cluster Sampling}
\label{sec:clu_smp}
Given the simulated clusters for each food class, we can start sampling images to simulate the personalized food consumption patterns. Based on the obtained food consumption pattern as illustrated in Section~\ref{sec:modified markov chain}, we know the appearance frequency for each food class, which allows us to learn the number of images we need to sample for each food class. In addition, we learn the user's preferred cooking style based on their provided initial food consumption pattern. Specifically, we calculate the standard deviation $\sigma$ of the initial food consumption pattern, which is used to construct a Gaussian distribution with the same $\sigma$ where a high $\sigma$ indicates that the person prefers a diverse range of cooking styles while a low $\sigma$ indicates that the person prefers a small range of cooking styles. Finally, given the appearance frequency and the constructed Gaussian distribution, we sample the appropriate number of food images from each cluster and randomly place them in the food consumption pattern to simulate the personalized food consumption pattern. 

% Given clusters in each food class, we need to sample clusters that will be used in food eating data patterns. In the simulated data pattern, we already know the number of times each food class appeared in the data pattern. Another piece of information we need is the eating habit of a person in terms of cooking style. Such information is obtained from the user's input as a standard deviation which is used for constructing a normal distribution later on. A normal distribution is used to know the probability of a cluster appearing in a food class. A high standard deviation indicates that the person prefers diverse cooking styles while a low standard deviation means the person prefer eating a small range of cooking style within a food class. Based on the built normal distribution, we can know how many images we need to sample in each cluster within a food class. The sampled images will be placed into the corresponding class in the data pattern but in random order.   

\section{Experiments}
\label{sec:experiments}
In this section, we conduct experiments to show that our modified Markov chain method can improve the quality of the simulated data pattern compared to the original Markov chain method or random simulation method. If the simulated data pattern correlates better with the initially provided data pattern, we know that the simulated data pattern is of higher quality. First, we manually select foods from the Food-101 dataset \cite{bossard14} as the initial data pattern. The image sampling process is also conducted on the Food-101 dataset \cite{bossard14}. We use Dynamic Time Warping distance \cite{Sakoe1978DynamicPA} and Kullback-Leibler Divergence \cite{shlens14} as our evaluation metrics. These two metrics can evaluate the correlation between the simulated pattern and the initial data pattern. Note that since there is no label for the future eating pattern, we cannot use the prediction accuracy directly as our metric. In addition, the goal of the simulation is not to make correct prediction. Instead, we want to simulate a food consumption pattern that correlates well with the initial data pattern so that a personalized classifier can learn from the data distribution. We use different lengths of the initially provided food consumption pattern in the experiments to see how length may affect performance. We also show examples of image sampling results on simulated food consumption for qualitative evaluation.

\subsection{Evaluation Metric}
\subsubsection{Dynamic Time Warping}
To evaluate how well the extended data pattern infers from the initially provided data pattern, we use Dynamic Time Warping (DTW) \cite{Sakoe1978DynamicPA} distance between the extended data pattern and the initially provided data pattern. Unlike Euclidean distance, which compares two data patterns by aligning only the corresponding points, DTW compares two data patterns by aligning one point in one data pattern to multiple points in another data pattern, which can better correlate the two data patterns. In general, DTW measures how well one sequence can follow the pattern of another sequence.

% We need a metric to evaluate whether the simulated pattern is good or not. The objective is to see how much of the extended data pattern is correctly inferred from the initially provided data pattern. In this case, we use Dynamic Time Warping (DTW) between the extended data pattern and the initially provided data pattern. This metric was proposed by H. Sakoe and S. Chib \cite{Sakoe1978DynamicPA} and C. A. Ratanamahatana and E. Keogh \cite{Ratanamahatana04} also did some review on this metric. Unlike the Euclidean distance metric, which compares two data patterns by aligning only the corresponding points, DTW compares two data patterns by aligning one point in one data pattern to multiple points in another data pattern, which maximally correlates the two data patterns.

\begin{comment}
\begin{figure}[htp]
    \centering
    \includegraphics[width=9cm, height=6cm]{DTW.png}
    \caption{Comparison of DTW distance and Euclidean distance}
    \label{fig:DTW}
\end{figure}
\end{comment}

In our case, DTW only needs to be applied on the Hamming distance since we just want to see whether the inference is correct or not instead of the degree of correctness. To calculate the DTW distance between time series $A$ and $B$ with lengths $m$ and $n$ respectively, we construct an $m \times n$ matrix. Each element of the matrix represents the distance between $A$ and $B$ at $(i,j)$. The distance is represented by $dis(A_i,B_j) = hamming(A_i,B_j)$. After constructing the matrix, we find the path with a minimum distance from $(0,0)$ to $(m,n)$ in the matrix. The cumulative distance is calculated using dynamic programming and can be represented by Equation ~\ref{eq:dtw}:

\begin{equation}
\label{eq:dtw}
\begin{split}
D_{dtw}(i,j) = dis(A_i,B_j) + min\{D_{dtw}(i-1,j-1),\\D_{dtw}(i,j-1),\\D_{dtw}(i-1,j)\}
\end{split}
\end{equation}

where $D_{dtw}$ represents the cumulative DTW distance at (i,j). A lower DTW distance indicates a better correlation between the simulated data pattern and the initial data pattern. However, only using DTW distance as our metric is not enough to confirm that the simulated food consumption pattern has been well inferred from the initially provided food consumption pattern. For example, The DTW distance may be small when a certain food consecutively appears multiple times in the simulated food consumption pattern since these consecutive decisions will only match one decision point in the initial data pattern and can cause a zero distance at that point, which makes total DTW distance very small. However, the simulated data pattern is likely not realistic. 
% The DTW distance may be small when consecutive repetitive occurrences of one food appears in one pattern since that pattern could always match only one point with the other pattern, but the appearance of each food class is very unbalanced in this pattern. 
This is particularly true for the original Markov chain method. Therefore, we need another metric to compare the performance of the original Markov Chain method and our modified version.  

\subsubsection{Kullback-Leibler Divergence}
Kullback-Leibler(KL) Divergence \cite{shlens14} is used to measure the degree of difference between two probability distributions. In our case, it measures the probability of each food appearing in the initial food consumption pattern and the simulated food consumption pattern and compares the difference between the two distributions. It is formulated as in Equation ~\ref{eq:kl}
\begin{equation} 
D_{KL}(p,q) = \sum_{i=1}^{n} p_i log(\frac{p_i}{q_i})
\label{eq:kl}
\end{equation}
where $p_i$ represents the probability of a food class appearing in the simulated food consumption pattern, $q_i$ represents the probability of a food class appearing in the initial food consumption pattern, and $n$ represents the number of food classes in the simulated food consumption pattern. A lower KL divergence indicates better probability distribution matching of food appearance between the simulated food consumption pattern and the initial food consumption pattern.

\subsection{Experiment Setup and Results}
Since our goal is to simulate a food consumption pattern that can be used for training a personalized classifier, the simulated data pattern needs to have some correlations with the initial data pattern. To check whether our simulated data pattern is successful, we randomly simulate a data pattern that has the same number of food classes as the data pattern simulated by the modified Markov Chain method. Then we compare the random simulation method with our modified Markov Chain method using DTW distance. We calculate the DTW distance and KL divergence on simulated food consumption patterns with different lengths of initial food consumption patterns and methods. We also use KL divergence to show that the modified Markov Chain method is better than the original Markov Chain method.

\subsubsection{Datasets}
To simulate a data pattern, we manually select food classes from the Food-101 \cite{bossard14} dataset to form initial food consumption patterns. Each initial data pattern can be used to form additional data patterns of different lengths to evaluate the effect of the initial data pattern length. The initial data pattern has a length of [5,10,20,30,40,50], and each length corresponds to 20 initial data patterns, giving us a total of 120 initial data patterns. We experimentally trained 20 sets for each length of the initial food consumption pattern. The image sampling is then evaluated based on the corresponding classes of images from the Food-101 dataset. 

\begin{figure*}[h]
    \centering
    \includegraphics[width=17cm, height=5.5cm]{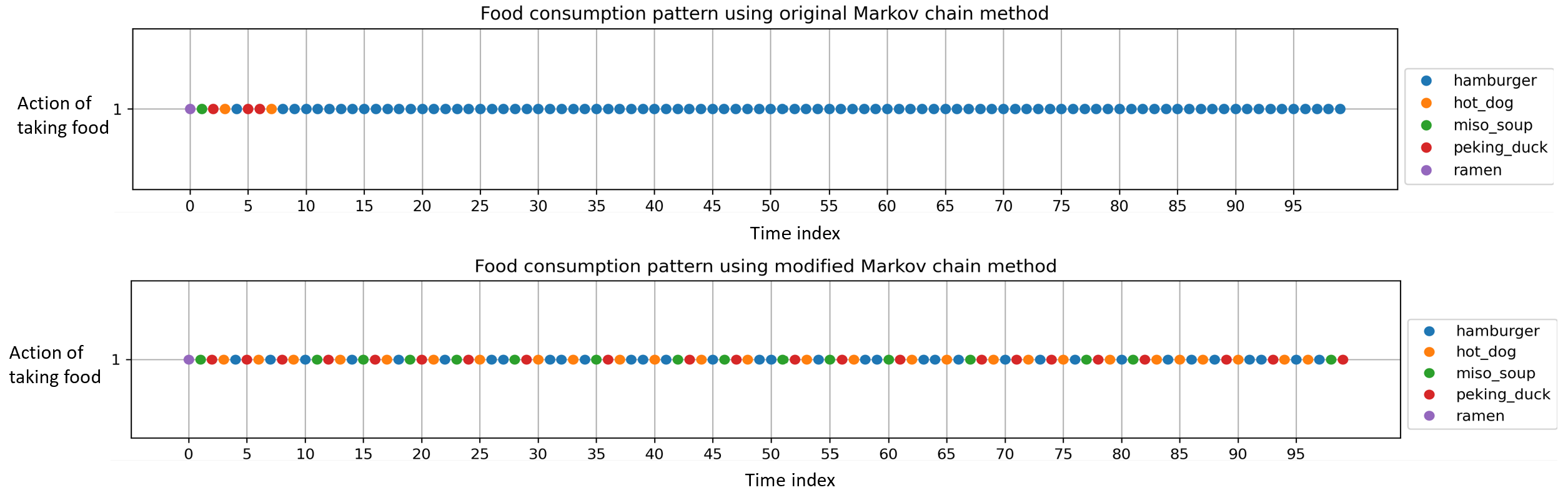}
    \caption{Comparison of food consumption pattern using original and modified Markov chain method}
    \label{fig:cmp_markov}
\end{figure*}

\subsubsection{Comparison between using the original and our modified Markov chain method}
As pointed out in Section~\ref{sec:dps}, there are some issues with using the original Markov chain method for our purpose. We conduct experiments to see the difference in the simulated food consumption pattern. The sample simulated food consumption is shown in Figure~\ref{fig:cmp_markov}. Different colored dots represent food class eaten at each unit of time assuming the simple case where only one type of food is consumed at a time. The first 5 units of time present the initially provided food consumption pattern. We can see the food consumption pattern simulated by the original Markov chain method always samples the same food type in later units of time, which is not realistic in real life. We also observe the food consumption pattern simulated by the modified Markov chain method better mimics the initial food consumption pattern and is more realistic. In Table~\ref{table:old1}, we use KL divergence to illustrate problems using the original Markov chain. We can see the KL divergence for the original Markov Chain method is large because there is a large difference between the probability of food appearing between the simulated data pattern and the initial data pattern. This issue can be addressed by the modified Markov chain method because if there is any consecutive occurrence of decisions, it will try to correct this bias. In addition, if a person prefers certain foods in the initial data pattern, the simulated food consumption pattern will still keep this preference. We can see from the results that the modified Markov chain method can obtain lower KL divergence, which shows a better matching to the pattern of food appearing in the initial data pattern.
% The simulated pattern seems to be more aligned with the initially provided pattern.
% \textcolor{red}{What about your modified method? You need to describe that too. Reply: just add the description.}

\begin{table*}[h]
\centering
\begin{tabular}{|p{2cm}|p{2.4cm}|p{2.4cm}|p{2.4cm}|p{2.4cm}|} 
 \hline
 Length of initial pattern & $Markov_{orig}$ & $Markov_{ours}$ & $Markov_{orig}$ with new & $Markov_{ours}$ with new\\ [0.5ex] 
 \hline\hline
 5 & $1.9\pm0.158$ & \textbf{0.531\boldsymbol{$\pm$}0.031} & $1.371\pm0.28 $ & \textbf{0.551\boldsymbol{$\pm$}0.12}\\ 
 10 & $1.35\pm0.54$ & \textbf{0.237\boldsymbol{$\pm$}0.176} & $0.957\pm0.371$ & \textbf{0.382\boldsymbol{$\pm$}0.16}\\
 20 & $1.07\pm0.34$ & \textbf{0.0756\boldsymbol{$\pm$}0.04} & $0.78\pm0.135$ & \textbf{0.301\boldsymbol{$\pm$}0.09}\\
 30 & $0.903\pm0.251$ & \textbf{0.085\boldsymbol{$\pm$}0.043} & $0.649\pm0.119$ & \textbf{0.269\boldsymbol{$\pm$}0.076}\\
 40 & $0.758\pm0.171$ & \textbf{0.123\boldsymbol{$\pm$}0.061} & $0.566\pm0.073$ & \textbf{0.282\boldsymbol{$\pm$}0.093}\\ 
 50 & $0.652\pm0.114$ & \textbf{0.138\boldsymbol{$\pm$}0.069} & $0.503\pm0.065$ & \textbf{0.238\boldsymbol{$\pm$}0.06}\\[1ex] 
 \hline
\end{tabular}
\caption{Comparison between the original Markov chain method and the modified Markov chain method using $ KL\; Divergence\pm standard\; deviation$ between the initial provided food consumption pattern and the simulated food consumption pattern (smaller value indicates better performance). Second and third columns show the results without adding new classes. The last two columns show the results with added new classes. \;$Markov_{orig}$ means original Markov chain method,\;$Markov_{ours}$ is our modified Markov chain method,\;$with\;new\;$ indicates with added new food classes}
\label{table:old1}
\end{table*}

\begin{table*}[h]
\centering
\begin{tabular}{|p{2cm}|p{2.4cm}|p{2.4cm}|p{2.4cm}|p{2.4cm}|} 
 \hline
 Length of initial data pattern & $Random$ & $Markov_{ours}$ & $Random$ with new & $Markov_{ours}$ with new\\ [0.5ex] 
 \hline\hline
 5 & $66.3\pm2.74$ & \textbf{55.6\boldsymbol{$\pm$}2.21} & $79.6\pm3.67$ & \textbf{57.15\boldsymbol{$\pm$}14.79} \\ 
 10 & $69.2\pm2.64$ & \textbf{64.95\boldsymbol{$\pm$}7.85} & $76.4\pm4.95$ & \textbf{63\boldsymbol{$\pm$}7.42}  \\
 20 &$67.75\pm3.73$ & \textbf{56.7\boldsymbol{$\pm$}2.85} & $73.5\pm2.78$ & \textbf{59.55\boldsymbol{$\pm$}5.32} \\
 30 & $68.4\pm3.7$ & \textbf{49.2\boldsymbol{$\pm$}5.62} & $73.95\pm2.64$ & \textbf{58.1\boldsymbol{$\pm$}6.54} \\
 40 & $68.25\pm3.04$ & \textbf{50.5\boldsymbol{$\pm$}4.96} & $72.85\pm3.41$ & \textbf{59.8\boldsymbol{$\pm$}6.15} \\ 
 50 & $71.6\pm2.5$ & \textbf{54.75\boldsymbol{$\pm$}5.52} & $75.1\pm3.48$ & \textbf{62.7\boldsymbol{$\pm$}4.27} \\[1ex] 
 \hline
\end{tabular}
\caption{Comparison between the original Markov chain method and the modified Markov chain method using $DTW\; Distance\pm standard\; deviation$ between the initially provided data pattern and the simulated data pattern with adding new food classes (smaller value indicates better performance). Second and third columns show the results without adding new classes. The last two columns show the results with added new classes. \;$Random$ means Random simulation method,\; $Markov_{ours}$ is our modified Markov chain method,\;$with\;new\;$ indicates with added new food classes}
\label{table:new1}
\end{table*}

\begin{figure*}[h]
    \centering
    \includegraphics[width=17cm, height=3.1cm]{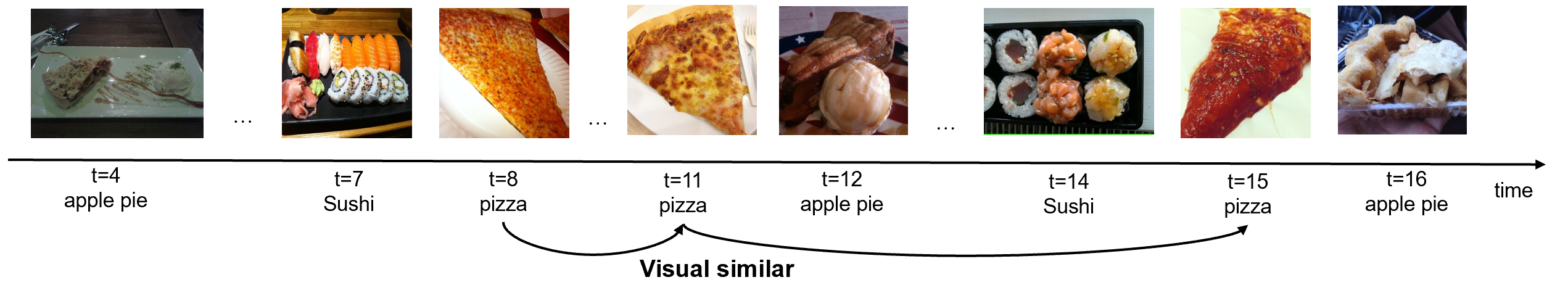}
    \caption{Image sampling over simulated data pattern}
    \label{fig:img_smp}
\end{figure*}

\subsubsection{Results Of Data Pattern Simulation}
The results of simulating food consumption patterns are shown in Table~\ref{table:old1}, and Table~\ref{table:new1}. In these tables, \boldsymbol{$Markov_{orig}$} means the original Markov chain method without adding new food classes during simulation. \boldsymbol{$Markov_{orig}\;with\;new$} means the original Markov Chain with added new food classes during simulation. \boldsymbol{$Markov_{ours}$} means our modified Markov chain method without adding new food classes. \boldsymbol{$Markov_{ours}\;with\;new$} means modified Markov Chain method with added new food classes during simulation. \boldsymbol{$Random$} means random simulation method without adding new classes. \boldsymbol{$Random\;with\;new$} means random simulation method with added new classes. The bold number in the table means the performance is better than another method.

\textbf{Results without adding new classes.} To show the results of simulating data patterns without adding new food classes on different methods, we ran experiments on 20 different initial data patterns. We then took the average DTW distance (for comparison between the random simulation method and modified Markov chain method) and average KL divergence(for comparison between the original Markov chain method and modified Markov chain method). We conduct experiments using initial data pattern lengths of [5,10,20,30,40,50]. In the second and third columns of Table~\ref{table:old1}, we observe higher KL divergence of the original Markov Chain method. Worse performance in the original Markov chain method is due to the unbalanced likelihood of food appearing in the simulated food consumption pattern. Therefore, the original Markov Chain method is not suitable to simulate the food consumption pattern. In the second and third columns of Table~\ref{table:new1}, we observe higher DTW distance of the random simulation method. Worse performance in the random simulation method is because the simulated pattern has almost no correlation with the initially provided pattern. The simulation does not follow the pattern in the initial data pattern. Finally, the modified Markov chain method can obtain a good performance when compared with both the original Markov chain method and random simulation method. The modified Markov chain method can simulate a pattern that has some correlation with the initial food consumption pattern. In addition, if the length of the initial food consumption pattern is large, the modified Markov chain can output competitive results in terms of KL divergence.

\textbf{Results with adding new classes.} We apply the same experiment setup when adding new classes. From the last two columns of Table~\ref{table:old1}, when new food classes are included in simulating future data patterns, we observe the KL divergence still decreases compared to the original Markov Chain method. From the last two columns of Table~\ref{table:new1}, we can observe the DTW distance also decreases compared to the random simulation method. Although the resulting values of using both metrics for comparing the methods are increased compared with the case without adding new classes, this is within expectation because in the simulated data pattern there are food classes that are not in the initial data pattern. 
% Adding new classes can affect the inference from the initial data pattern. The probability of each food appearing can be affected because of the inclusion of new classes in the simulated data pattern while these new classes are not in the initial data pattern. 
With longer initial data patterns, the KL divergence is lower for the modified Markov chain method as shown in Table~\ref{table:old1}. However, the DTW distance does not decrease in this case as shown in Table~\ref{table:new1} because the number of new food classes included in the simulated pattern is increasing.

% \textcolor{red}{Discussion of Table 1 should go here. Also, what about the comparison between the original Markov chain and your modified version? Reply: Comparison just added. New metric has been applied.}

\subsubsection{Image sampling over simulated food consumption pattern}
Figure~\ref{fig:img_smp} shows the sample results of image sampling over a simulated data pattern. We select sample images of apple pie, pizza, and sushi from the Food-101 dataset. We notice that image samples for the same food category appear visually similar in the data pattern at different time steps, which aligns with our assumption that a person typically prefers the same cooking style resulting in a visually similar appearance of the foods over time.
%\textcolor{red}{Can we provide some quantitative measure of the visual similarity? Also, maybe comparison to not using self-supervised learning results?}

\begin{comment}
\subsection{Discussion Of Results}
Our results show that using the modified Markov chain method to simulate an eating data pattern is successful. Generally, in the case of not adding new classes, the longer the initial data pattern which means more training data provided, the smaller the DTW distance and KL divergence between the initially provided data pattern and the simulated data pattern. However, we can see that this effect is not as significant when the length of the initial data pattern is very large. 

When we add new classes, the overall performance is worse than when we do not add new classes. This is within expectations because of the increased randomness when adding new classes. Because the overall relative weight becomes smaller as the length of initial data pattern becomes smaller, we can still say the experiment is successful. However, we can also see that the longer initial data pattern cannot make the DTW distance smaller in this case. 

Overall, our method of simulating an eating data pattern is convincing. However, very long initial data patterns cannot help much in simulating a data pattern because of added noise. If we can control an appropriate initial data length, we can simulate a data pattern that matches a person's eating habits.
\end{comment}

\section{Conclusion}
%In this paper, we have shown the benefits of simulating a food eating data pattern, which can make the research in personalized classifiers more effective. 
In this paper, we proposed to simulate food consumption pattern using a modified Markov chain method. Our method can accommodate new foods not included in the initial data pattern and closely mimic user preference of food choices and their occurrence as provided in the initial data pattern. Experimental results show that our proposed method produces realistic data pattern compared to using the original Markov chain method and a random simulation method. 

% An efficient and accurate simulated food consumption pattern is highly beneficial for future research in personalized classifiers. In this paper, we simulated the food consumption data pattern using a modified Markov chain method, considering other factors such as new foods, and preferences shown by the participant to make the simulated data pattern more realistic. Our experimental results prove that our method of simulating a food consumption data pattern is better than the original Markov chain method and random simulation method, which shows our method is successful. 

Our future work will focus on improving the simulation of food consumption patterns incorporating other aspects. For example, we can consider the timing of food consumption such as breakfast, lunch, and dinner. We can use a hierarchical classification of food types and use such hierarchy to simulate what someone may eat during a meal. Methods to simulate food consumption patterns efficiently and accurately could benefit the development of personalized classifiers to improve food image classification.
% We believe our research has a large potential in the future and can be beneficial to image-based dietary assessment.
\begin{acks}
This work was supported by the National Institutes of Health under Grant 1U24CA268228-01.
\end{acks}

\bibliographystyle{ACM-Reference-Format}
\bibliography{sample-base}

\end{document}